\pdfoutput=1
\documentclass[conference]{IEEEtran}
\usepackage{cite}
\usepackage{amsmath,amssymb,amsfonts}
\usepackage{algorithmic}
\usepackage{graphicx}
\usepackage{textcomp}
\usepackage{xcolor}
\def\BibTeX{{\rm B\kern-.05em{\sc i\kern-.025em b}\kern-.08em
    T\kern-.1667em\lower.7ex\hbox{E}\kern-.125emX}}

\usepackage{cite}
\usepackage{amsmath,amssymb,amsfonts}
\usepackage{booktabs}
\usepackage{algorithmic}
\usepackage{graphicx}
\usepackage[pdftex,colorlinks=true,pdfstartview=FitV,bookmarks=false]{hyperref}

\usepackage{gensymb}
\usepackage[export]{adjustbox}
\usepackage{textcomp}
\usepackage{makecell}
\usepackage{arydshln}
\usepackage{xcolor}
\usepackage{svg}
\usepackage{pdfpages}
\usepackage{float}

\newcommand{\mycomment}[1]{}

\begin{document}

\title{Language-Based Augmentation to Address Shortcut Learning in Object-Goal Navigation\\}

\author{\IEEEauthorblockN{1\textsuperscript{st} Dennis Hoftijzer}
\IEEEauthorblockA{
\textit{University of Twente}\\
Enschede, The Netherlands \\
dennishoftijzer@gmail.com}
\and
\IEEEauthorblockN{2\textsuperscript{nd} Gertjan Burghouts}
\IEEEauthorblockA{
\textit{TNO}\\
Den Haag, The Netherlands \\
gertjan.burghouts@tno.nl}
\and
\IEEEauthorblockN{3\textsuperscript{rd} Luuk Spreeuwers}
\IEEEauthorblockA{
\textit{University of Twente}\\
Enschede, The Netherlands \\
l.j.spreeuwers@utwente.nl}
}

\maketitle

\begin{abstract}
Deep Reinforcement Learning (DRL) has shown great potential in enabling robots to find certain objects (e.g., `find a fridge') in environments like homes or schools. This task is known as \textit{Object-Goal Navigation} (ObjectNav). DRL methods are predominantly trained and evaluated using environment simulators. Although DRL has shown impressive results, the simulators may be biased or limited. This creates a risk of \textit{shortcut learning}, i.e., learning a policy tailored to specific visual details of training environments. We aim to deepen our understanding of shortcut learning in ObjectNav, its implications and propose a solution. We design an experiment for inserting a shortcut bias in the appearance of training environments. As a proof-of-concept, we associate room types to specific wall colors (e.g., bedrooms with green walls), and observe poor generalization of a state-of-the-art (SOTA) ObjectNav method to environments where this is not the case (e.g., bedrooms with blue walls). We find that shortcut learning is the root cause: the agent learns to navigate to target objects, by simply searching for the associated wall color of the target object's room. To solve this, we propose \textit{Language-Based} (L-B) \textit{augmentation}. Our key insight is that we can leverage the multimodal feature space of a Vision-Language Model (VLM) to augment visual representations directly at the feature-level, requiring no changes to the simulator, and only an addition of one layer to the model. Where the SOTA ObjectNav method's success rate drops 69\%, our proposal has only a drop of 23\%. Code is available at \url{https://github.com/Dennishoftijzer/L-B_Augmentation} \\
\end{abstract}


\section{Introduction}
\begin{figure}[tbp]
    \centering
     \includegraphics[trim=11cm 2.6cm 12cm 10cm, clip, width=0.49\textwidth]{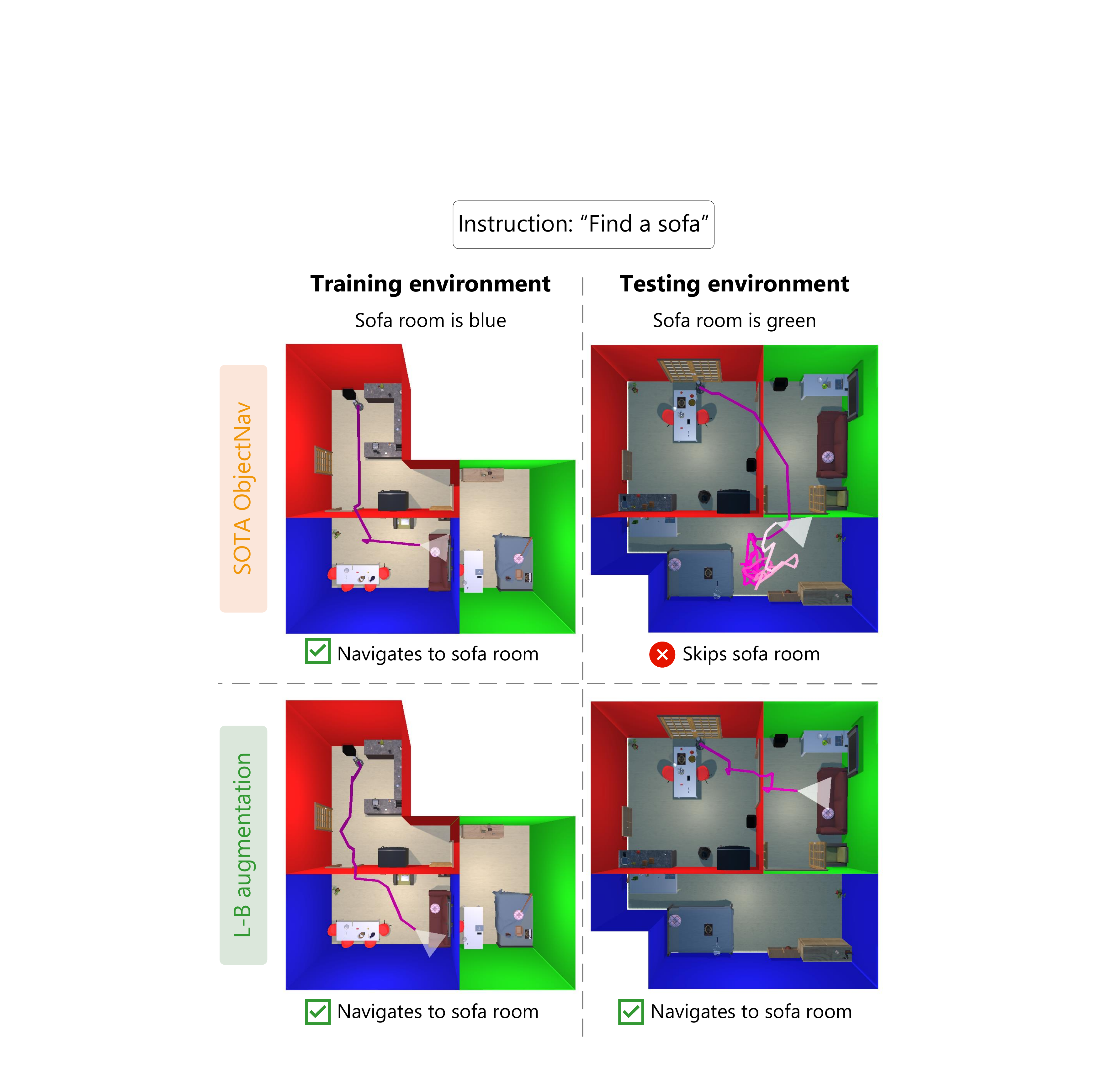}
     \caption{We propose \textbf{Language-Based (L-B) augmentation} to generalize better to scenes with different wall colors. In this example, we interchange the wall color of the bedroom and living room, causing the SOTA objectNav method \cite{EmbCLIP} to look for the sofa in the blue bedroom (wrong). With our augmentations, this is countered.}
     \label{fig:intro_fig}
 \end{figure}

Humans can easily find certain objects (e.g. `find a fridge') in complex environments we have not seen before, such as a friend's house. We effortlessly avoid any obstacles but also reason about the unseen environment to decide which room to explore next e.g., `where is a fridge most likely located?'. The embodied-AI (E-AI) community has made great strides by learning embodied agents (or `virtual robot') such skills using Deep Reinforcement Learning (DRL) in a task called Object-Goal Navigation (ObjectNav) \cite{Objectnav_revisited}. Despite good progress, DRL relies on gathering experience over millions (or billions) of iterations, making it impossible to learn in real-world environments. Therefore, research has been drawn to E-AI simulators \cite{AI2THOR, Habitat, RoboTHOR,iGibson_simulator}, which allow for easily training agents in various simulated 3D indoor environments. However, subtle, but detrimental, dataset biases might arise in E-AI simulators due to data collection artifacts, limitations in rendering, or simply unintended biases the simulator designer is not aware of. For instance, all kitchens in training environments might have a tiled floor. Consequently, training in E-AI simulators creates a profound risk of \textit{shortcut learning} \cite{shortcut_learning}: learning a simple, non-essential policy, tailored to specific details of the simulated environment, rather than learning any semantic reasoning or task-related skills. Efficient object-goal navigation involves learning useful semantic priors such as object-room relations (e.g., a fridge is in the kitchen), however can easily lead to unintended shortcuts (e.g., fridge is located near a tiled floor), which fail to generalize to environments where the shortcuts are no longer valid.


In this work, we deepen our understanding of shortcut learning in ObjectNav, its implications and propose a solution. First, we introduce an \textit{out-of-distribution} (o.o.d.) generalization test. We insert a dataset bias in the appearance of training environments, which offers the agent a shortcut pathway for finding a given target object. As a proof-of-concept of such a shortcut bias, we associate each room type to a unique wall color i.e., kitchens have red walls, bedrooms have green walls and so forth. Using our setup, we are able to evaluate o.o.d. generalization of a state-of-the-art (SOTA) ObjectNav method \cite{EmbCLIP} to environment where we change wall colors (e.g. kitchens have blue walls). As a result, we find that (1) only changing wall colors degrades performance significantly, and (2) shortcut learning is the root cause. The agent learns to navigate towards target objects by simply searching for the wall color associated with the target object’s room.

Secondly, we aim to increase domain generalization. Domain randomization methods e.g., randomizing textures, colors and shapes of objects or environments are commonly used to transfer policies in DRL \cite{domain_rand_robotics, Domain_rand_flight, ProcTHOR}. However, these methods specifically require changes to the simulator, which might be inflexible or difficult to modify e.g., high-fidelity simulators with training data reconstructed from real-world 3D scans \cite{MP3D, HM3D}. While more sophisticated methods for partially editing individual frames during training exist (e.g., text-to-image models \cite{stable_diffusion, DALLE}), they are slow, computationally expensive and error-prone. Instead, we take a different approach and propose \textit{Language-Based (L-B) augmentation} (see Fig. \ref{fig:intro_fig}). We augment directly at feature-level, without editing individual frames or any changes to the simulator.

We build upon promising results from \cite{EmbCLIP}, where visual representations within the agent's architecture are based on a Vision-Language Model (VLM). RGB observations are encoded using a Contrastive Language Image Pretraining (CLIP) \cite{CLIP} visual backbone. CLIP jointly trains an image and text encoder, such that both produce similar representations for visual concepts in images or their names in natural language. Our key insight is that we can augment agent's visual representations at feature-level, by describing variations of the dataset bias in natural language. By an elegant modification to the SOTA architecture \cite{EmbCLIP}, with only one additional layer, we generalize better to environments with different wall colors in ObjectNav.


\section{Related work}
\subsection{Vision-Language for visual navigation}
Several recent works have proposed utilizing pre-trained features of Vision-Language Models (VLMs), pre-trained on internet-scale data, for several visual navigation tasks \cite{CLIP_On_Wheels2022, LMNAV, huang2022visual, ZSON, EmbCLIP}. In \cite{EmbCLIP}, authors explore the effectiveness of learning a navigation policy based on CLIP embeddings \cite{CLIP}. With their method, EmbCLIP \cite{EmbCLIP}, they show that CLIP's visual representations encode useful navigation primitives such as reachability and object localization. They set new SOTA results on several visual navigation tasks, including ObjectNav, and show promising results for generalizing to an open-world setting i.e., navigating to target objects not seen during training. Moreover, as generating the robot trajectories and paired language annotations in the real world might be costly, further works have proposed utilizing VLMs out-of-the-box with maps to enable zero-shot navigation i.e., without supervision of DRL reward signals or human demonstrations \cite{CLIP_On_Wheels2022, LMNAV, huang2022visual}. We adopt the architecture of EmbCLIP as a baseline, given its strong performance on a variety of settings. However, our focus is specifically on addressing shortcut bias in E-AI simulators, which might impede generalization of DRL methods to novel environments.


\subsection{Embodied AI simulators and scene datasets}
Many E-AI simulators  \cite{AI2THOR, Habitat, RoboTHOR,iGibson_simulator} have been developed, along with several (near) photo-realistic scene datasets \cite{Gibson_dataset, MP3D, HM3D, ProcTHOR}. Scenes can be either reconstructed from 3D scans of real-world houses e.g., Matterport \cite{MP3D}, or synthetically composed from artist created 3D assets e.g., RoboTHOR \cite{RoboTHOR}. Both methods are extremely costly to collect. Reconstructing scenes from 3D scans involves stitching images from specialized cameras whilst manually composing synthetic scenes involves carefully configuring lighting, object placement and textures. ProcTHOR \cite{ProcTHOR} recognizes this fact and instead uses a procedural generation process to generate 10,000 scenes (dubbed ProcTHOR-10k). In this work, we leverage the ProcTHOR-10k scene dataset. Due to the procedural generation process, we can fully customize these scenes by altering the appearance of individual objects and room surfaces (walls, floors and ceilings). For instance, a red sofa can be replaced with a black one. This customization and our proposed interventions on ProcTHOR-10k, allow for inserting a shortcut bias in the appearance of training scenes and evaluate o.o.d. generalization. 

Although ProcTHOR enables E-AI to scale, this does not imply shortcut biases completely disappear. Recent works show that even Large-Language Models (LLMs), pre-trained on text amounting to billions of words, suffer from shortcut learning, largely due to collection artifacts in training data \cite{shortcut_learning, LLMshortcut}. Moreover, shortcut bias might be difficult to observe for humans e.g., superficial statistics in training data such as textures of specific frequencies in image classification tasks \cite{Wang_2023_ICCV}. Consequently, in E-AI simulators, shortcut bias might arise inadvertently. For instance, as ProcTHOR is generated procedurally, some smaller objects (e.g. a pen) are always placed on larger objects (e.g. a desk in the bedroom). An agent might learn a bias for navigating to a target object only when it is placed on this larger object and not when the object is placed independently (e.g. on the floor in the living room). We employ a simple wall color bias as a proof-of-concept for such unintended shortcut biases.

\subsection{Shortcut learning}
\textit{Shortcut learning} is emerging as a key impediment in the generalization ability of deep neural networks (DNNs) \cite{shortcut_learning}. Shortcuts are decision rules, often learned by DNNs, which aid performance on a particular dataset but do not match with human-intended ones. Accordingly, they typically fail when tested in only slightly different conditions. Prior work in shortcut learning is predominantly concerned with supervised learning \cite{shortcut_cues, shortcut_imagenet, Wang_2023_ICCV}. Similar to our work, \cite{shortcut_cues} designs an experimental setup to observe whether DNNs prefer to adopt color, shape or size shortcuts, and find DNNs naturally prefer certain shortcuts. In contrast, we study the shortcut learning phenomenon in the context of DRL. 

A common implication of shortcut learning in DRL is observed when transferring policies from simulation to the real-world \cite{shortcut_learning, domain_rand_robotics, splitnet}. Most policies trained in simulation generalize poorly to the real-world due to agents adapting to specific visual details of the simulator. Prior works cope with this so-called `reality gap' by domain randomization methods i.e., randomizing appearances in training environments \cite{domain_rand_robotics, Domain_rand_flight}. Similarly, ProcTHOR \cite{ProcTHOR} allows for randomizing e.g., textures and colors of walls, ceilings, floors and objects. While ProcTHOR shows incredibly powerful results, such augmentations might not be available for all simulators, and more often than not, difficult to apply post-hoc. Contrary, our method can readily be applied post-hoc as it requires no changes to training data or the simulator. We propose augmentations where we use targeted randomization of specific unintended biases, in our case, wall color. Although a simple wall color bias might be addressed using conventional domain randomization, these methods are inconvenient considering more intricate biases (e.g., a pen is always on a desk). In contrast, our method utilizes free-form natural language, which allows for easily adapting to different biases. Vision-Language Models (VLMs) e.g., CLIP \cite{CLIP}, allows us to augment at feature-level based on prior knowledge of the environment without any changes to training data.

\mycomment{
\begin{figure}[h!]
     \centering
     \includegraphics[trim=4.3cm 5.7cm 4.4cm 5.6cm, clip, width=0.3\textwidth]{Figures/Method_visually_identical.pdf}
     \caption{\textbf{Example house after proposed interventions.} Here, we show a 3-room house selected from the ProcTHOR 10k train split, where we select 3 target object categories per room type e.g., a sofa and television in the living room, a bed and dresser in the bed room, and a kettle and fridge in the kitchen.} 
     \label{fig:visually_identical}
 \end{figure}
}

\begin{figure*}[!h]
    \centering
     \includegraphics[trim=3cm 3.4cm 11cm 2.8cm, clip, width=\textwidth]{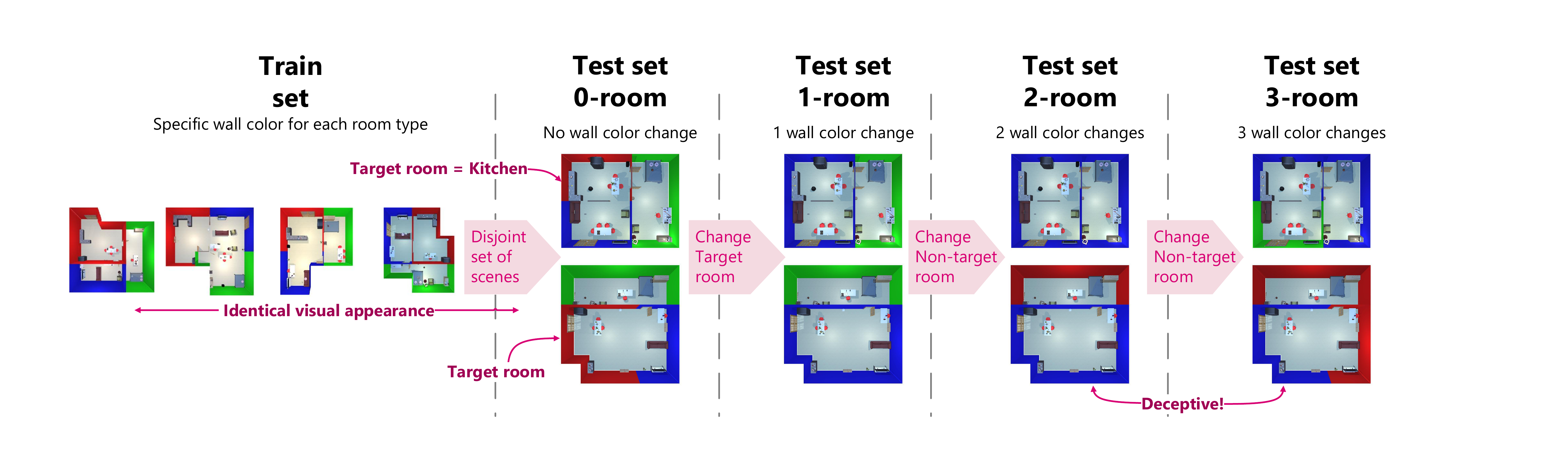}
     \caption{\textbf{Setup for our o.o.d. generalization test.} In this example, the target room is the kitchen (red walls in test set 0-room). We change the target room first (test set 1-room) and incrementally change more rooms (test set 2/3-room). The bottom row shows two examples of deceptive changes, where the wall color associated with the target room (red wall color) is moved to a different room type. The top row only shows nondeceptive changes.}
     \label{fig:dataset_gen}
 \end{figure*}

\section{ObjectNav preliminaries}
\subsection{ObjectNav definition}
In ObjectNav \cite{Objectnav_revisited}, agents are initialized at a random pose in an unseen environment and given a label specifying the target object category (e.g. `Bed'). The goal for the agent is to navigate to an instance of the target object within a certain time budget ($T=500$). At each time step $t$, the agent receives an image from an RGB forward-facing camera and can take one of 6 possible actions: move ahead, rotate left, rotate right, look up, look down and done. We do not utilize any depth sensor readings. Also, we simulate actuation noise to better resemble actuation in the real-world. A full description of the discrete action space is shown in Table \ref{tab:action_space} (Appendix). 

An episode is considered successful if (1) the agent executes the done action; (2) The target object is within a certain distance threshold, typically $d_t = 1$m; and (3) the target object is considered visible i.e., within the camera’s field of view and not fully obstructed.

\subsection{Evaluation metrics}
Following standard ObjectNav procedure \cite{Objectnav_revisited}, we report two primary evaluation metrics: the average success rate over all $N$ evaluation episodes (Success) and Success weighted by normalized inverse Path Length (SPL) \cite{Objectnav_revisited}, a measure for path efficiency. SPL is bounded by $[0,1]$, where $1$ is optimal performance i.e., the agent took the shortest path possible in all $N$ evaluation episodes. Note that, SPL is a stringent measure. Achieving an SPL of 1 is infeasible (even for humans), without knowing the target object location a priori. Additionally, we report two more evaluation metrics: Distance To Target (DTT) and the episode length. DTT is the remaining shortest path length to visibly see the target object. 

\section{O.o.d. test: interventions on ProcTHOR-10k}
\label{meth:interventions_on_ProcTHOR}
\subsection{Interventions on ProcTHOR-10k}
In order to evaluate o.o.d. generalization, we need to guarantee only to measure performance degradation due to changing wall colors, without other aspects (object appearances, scene layout, etc.) influencing our evaluation. Therefore, we propose some interventions on ProcTHOR-10k. First, we start by selecting a more uniform subset of scenes and targets. We select only houses with 3 rooms, which all consist of 3 room types: kitchen, bedroom and living room. For each room type, we select 3 target object categories (9 total) which are semantically related (e.g. fridge in kitchen). Section \ref{app:target_selection} (Appendix) shows an overview of all target objects selected. Next, we restrict ourselves to scenes which contain exactly one instance of each target object category in the associated target room e.g., every house contains 1 bed, positioned in the bedroom. We ensure this restriction by (1) selecting scenes which contain at least one instance of each target object in the associated room type and (2) manually removing any double (or more) instances of target objects. Secondly, we set identical appearances for all object categories (including doors) e.g., all chairs appear exactly alike, and identical appearances of room surfaces for each room type e.g., all kitchens have identically colored walls, floors and ceilings. We set object appearances identical by assigning one 3D asset from the ProcTHOR library to each object type. We set all room surfaces identical by setting the same materials from the ProcTHOR library. Lastly, we remove windows and wall decoration. These interventions limit the house variations to just wall colors, which is the only aspect influencing the performance. 


\subsection{Evaluating o.o.d. generalization}
We aim to assess the agent's generalization across increasing numbers of changing wall colors (illustrated in Fig. \ref{fig:dataset_gen}). The training set consists of visually identical houses, where living rooms have blue walls, kitchens have red walls and bedrooms have green walls. For our test sets, we use houses with a different layout such that agents cannot simply memorize object locations, and permute wall colors. The 0-room test set serves as a reference. To solely evaluate generalization to different wall colors, we use the same layouts in each test set and compare performance to the reference 0-room test set. First, we change the wall color of the target object's room as this is the simplest deviation (1 wall color change). For instance, if the target object is a fridge, we start by altering the wall color of the kitchen from red to e.g., green, whereas if the target object is a bed, we start with changing the wall color of the bedroom from green to e.g., blue. Next, we change another room's wall color (2 wall color changes). Finally, we change the wall colors of all three rooms (3 wall color changes). We change to all possible permutations (e.g. red kitchen to blue and green wall colors in test set 1-room) with repetition i.e., multiple room types can have the same wall color.


We differentiate wall color changes of non-target rooms (test set 2- and 3-room) into two types: \textit{deceptive} vs \textit{nondeceptive}. We expect that moving the learned color i.e, the wall color associated with the target room, to a non-target room will have a high impact, because the agent may look in the latter, wrong room. We refer to this wall color change as `deceptive'. Examples of deceptive changes are shown in the bottom row (test set 2- and 3-room). Instead, when none of the rooms has the learned color, we expect less performance degradation, because the agent is not misled. We refer to such a wall color change as `nondeceptive' (top row). 

\section{Method: Language-based Augmentation}
\label{meth:LB_aug}

\begin{figure*}[!h] 
    \centering
     \includegraphics[trim=0.6cm 6.9cm 5.1cm 3.3cm, clip, width=0.9\textwidth]{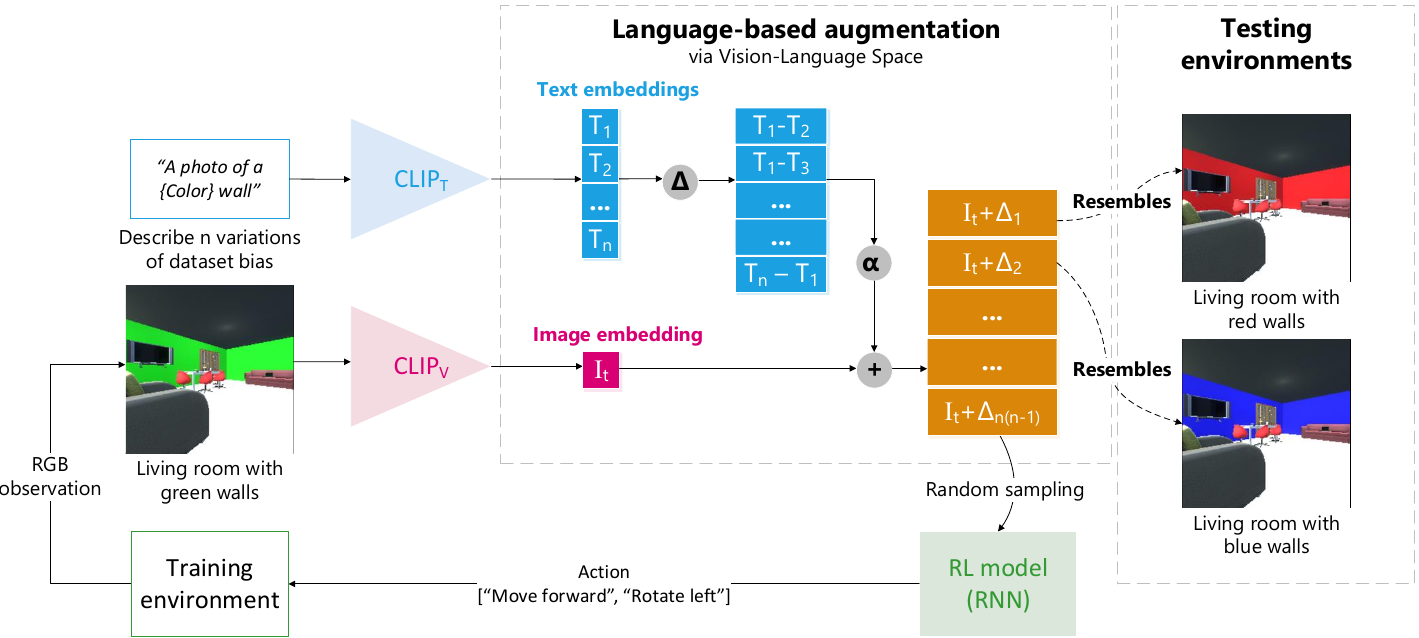}
     \caption{\textbf{Language-Based (L-B) augmentation via a the feature space of a vision-language space.} Our key insight is that we can augment agent's visual representations ($I_t$) using differences ($\Delta$) between encoded text descriptions of variations of the dataset bias ($T_{1,...,n}$). The augmented embedding of an image `A living room with green walls' resembles a `living room with red or blue walls'. The RL model (RNN) is not able to use a shortcut strategy even if during training living rooms always have green walls.}
     \label{fig:VL_augmentation}
 \end{figure*}

We increase domain generalization by augmenting the agent's visual representations at feature-level, such that these are more invariant to changing environments. We implement this by adding one layer on top of EmbCLIP \cite{EmbCLIP}. In EmbCLIP, the agent's visual representations are based on a VLM (CLIP). We leverage the vision-language representations for feature-level augmentations, without the need to modify the simulator. Our augmentations are based on textual descriptions of variations of the dataset bias that we want the agent to learn and generalize. We call this \textit{Language-Based (L-B) augmentation} (Fig. \ref{fig:VL_augmentation}). In EmbCLIP, at each time step $t$, a visual representation or image embedding $\pmb{I_t}$ is obtained by encoding RGB observations using CLIP's \cite{CLIP} visual encoder (CLIP$_v$). CLIP learns to associate text strings with their visual concepts in images. Our key insight is that we can represent domain specific knowledge, regarding the changes in environment appearances, using natural language. By encoding text descriptions of variations of the dataset bias (e.g. `a blue wall'), using CLIP's text encoder ($\text{CLIP}_T$), we vary visual representations without actually having seen images containing these variations (e.g. an image of a blue wall). This allows us to augment directly at feature-level. For encoding the text descriptions we use the default prompt template recommended by \cite{CLIP}: `a photo of a \{label\}'. We insert descriptions of variations of the dataset bias (e.g. `red wall'). We obtain our augmented embeddings $\pmb{I^{LB}_t}$ by computing differences between $n$ encoded text descriptions of variations of the dataset bias $\pmb{T_{1,...,n}}$, and adding to visual representation $\pmb{I_t}$:

\begin{align}
    \pmb{I^{LB}_t} &= \pmb{I_t} + \alpha \cdot \Delta(\pmb{T}), \label{eq:aug} \\
    \Delta(\pmb{T}) &= \begin{bmatrix} \pmb{\Delta_1} \\ \pmb{\Delta_2} \\ \vdots \\ \pmb{\Delta_{n(n-1)}} \end{bmatrix} = \begin{bmatrix} \pmb{T_1} - \pmb{T_2} \\ \pmb{T_1} - \pmb{T_3} \\ \vdots \\ \pmb{T_n} - \pmb{T_1} \end{bmatrix}
    \label{eq:text_embed}
\end{align}

where, $\alpha$ controls the degree of augmentation and $\Delta$ computes differences of all permutations of length 2 of text descriptions $\pmb{T}$. By randomly sampling an augmented embedding from $\pmb{I^{LB}_t}$ at each time step, we aim to provide the RL model (RNN) with an embedding which resembles the same room type (e.g., a living room in Fig. \ref{fig:VL_augmentation}), but with a different wall color (e.g., red and blue instead of green in Fig. \ref{fig:VL_augmentation}). We empirically find $\alpha = 50$ to work well by tuning for our specific dataset and shortcut bias. We standardize features before feeding into the RNN to ensure stability during training (as some features might dominate the loss due to large norms). In our case, we insert three ($n=3$) text descriptions of variations of the dataset bias: `blue wall', `red wall' and `green wall'. This results in $6$ augmented embeddings $\pmb{I_t}+\pmb{\Delta_{n(n-1)}}$ per image embedding $\pmb{I_t}$.

\section{Experiments}
We perform two experiments, where we aim to: (1) evaluate generalization of EmbCLIP \cite{EmbCLIP} to scenes where we change wall colors and study to what extend shortcuts influence the generalization ability; and (2) validate L-B augmentation can increase domain generalization by integrating within the architecture of EmbCLIP.

\subsection{Experimental setup}
\subsubsection{ObjectNav dataset details}
We train agents on 20 visually identical but biased scenes, generated using our proposed interventions (Section \ref{meth:interventions_on_ProcTHOR}). During training, we randomly sample 1 of 9 target objects. We analyze o.o.d. performance on a set of 5 disjoint scene layouts. We run 1080 evaluation episodes per test set to evenly distribute episodes over the 5 layouts, possible wall color permutations and target objects. Section \ref{App:eps_split} (Appendix) details the distribution.


\subsubsection{Agent architecture and configuration}
We use the ObjectNav EmbCLIP architecture \cite{EmbCLIP}, which has two different variations: a closed-world architecture, which assumes known target objects, and an open-world variant. Both encode RGB egocentric views using a frozen CLIP image encoder with a ResNet-50 backbone. However, the closed-world variant obtains a goal-conditioned embedding before feeding into an RNN, which involves removing the final layers from CLIP, whereas the open-world variant feeds the image embedding directly. For our o.o.d. generalization test, we adopt EmbCLIP's closed-world architecture given its better performance in a closed-world setting. We integrate our L-B augmentations in the open-world variant, as this architecture feeds the CLIP image embedding directly into an RNN, which allows for substituting this image embedding with a L-B augmented embedding, using random sampling. Following typical ObjectNav setup \cite{ProcTHOR, EmbCLIP}, the embodied agent approximately matches a LoCoBot with a $90\degree$ horizontal camera field of view, a $0.25m$ step size and a $30\degree$ turn angle.
 
\subsubsection{Reward setting}
At each time step $t$, the reward $r_t$ is: 
\begin{align}
    r_t= max(0,min\Delta_{0:t-1} - \Delta_t) + r_{slack} + r_{succ}
\end{align}
where $min\Delta_{0:t-1}$ is the minimal path length from the agent to the target object that the agent has previously observed during the episode, $\Delta_t$ is the current path length, $r_{slack} = -0.01$ is the slack penalty and $r_{succ} = 10$ is a large reward for successful episodes. The reward is shaped to optimize path efficiency and, therefore, SPL. 

\mycomment{
where:
\begin{itemize}
    \item $min\Delta_{0:t-1}$ is the minimal path length from the agent to the target object that the agent has previously observed.
    \item $\Delta_t$ is the current path length from the agent to the target.
    \item $r_{slack} = -0.01$ is the slack penalty.
    \item $r_{succ} = 10$ is a large reward for successful episodes.
\end{itemize}
}

\subsubsection{Implementation details}
We use the Allenact framework \cite{allenact} and render frames at $224 \times 224$ resolution. To parallelize training, we use DD-PPO \cite{DDPPO_PointNav} with 40 environment instances. After each rollout, the model is updated using 4 epochs of PPO \cite{GAE} in a single global batch size of 7680 frames. We perform validation every 200,000 frames and report results of the checkpoint with highest SPL. Additional hyperparemeters are shown in Section \ref{App:train_hyperparams} (Appendix).  

\subsection{Impact of Shortcut Learning}
\label{exp:Objectnav_generalization}

\begin{figure*}[!h]
     \centering
     \includegraphics[trim=-0.1cm 0.5cm -0.1cm -0.5cm, clip, width=\textwidth]{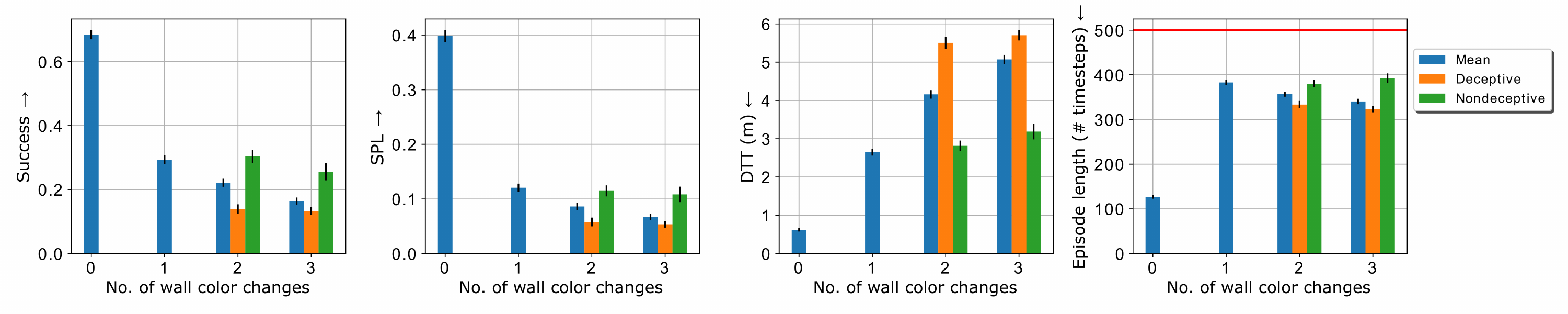}
     \caption{\textbf{Degradation for o.o.d. cases.} Performance of EmbCLIP \cite{EmbCLIP} to scenes with different wall colors. When only changing the wall color of the target object's room (1 wall color change), we already observe a large decrease in performance in all metrics.}
     \label{fig:res_ood_gen_test}
 \end{figure*}

\begin{figure}[tbp]
     \centering
     \includegraphics[trim=13.5cm 27.5cm 13.8cm 22.6cm, clip, width=0.49\textwidth]{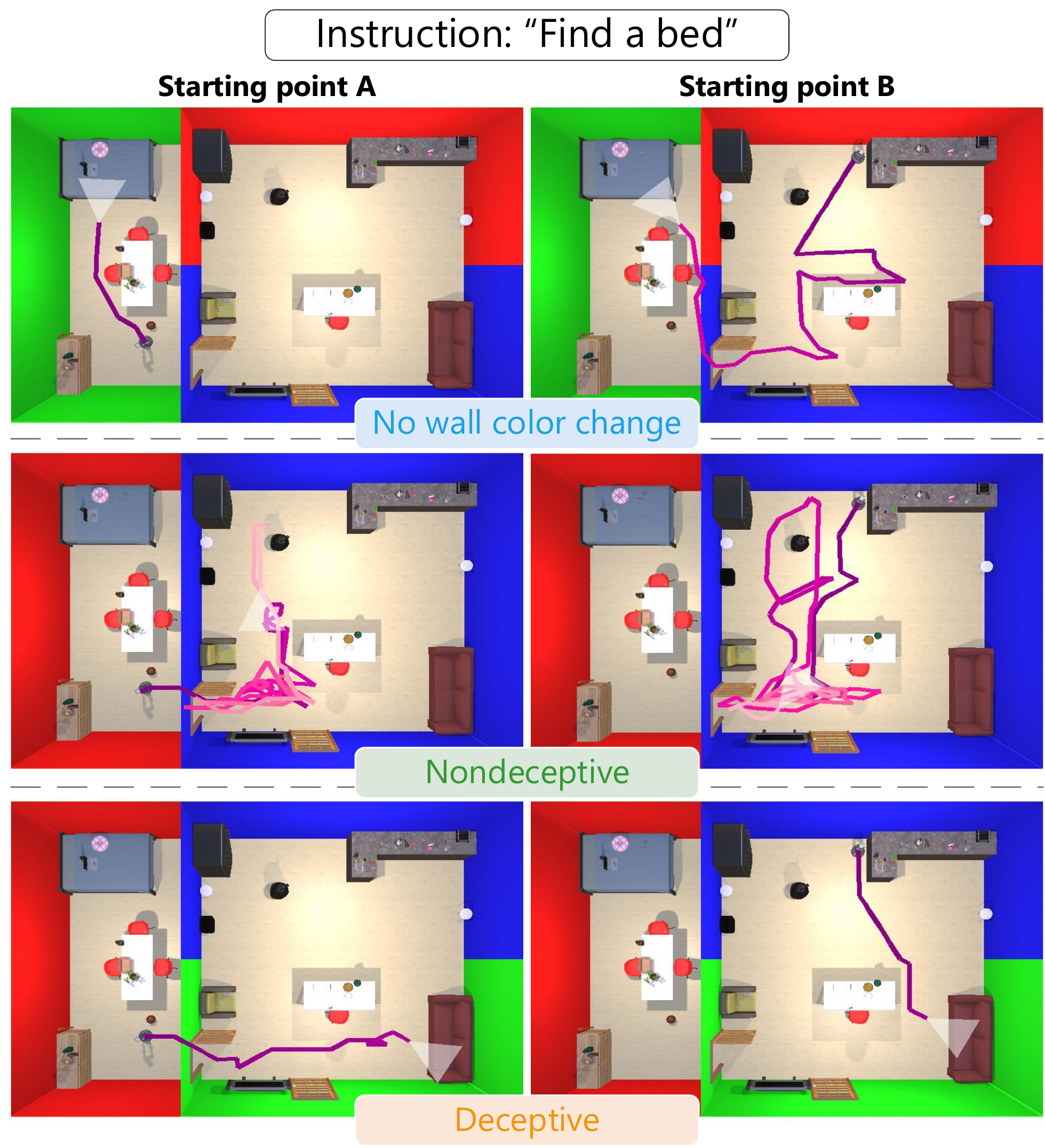}
     \caption{\textbf{Errors and shortcuts by the SOTA ObjectNav method.} We show example trajectories from 2 different starting position (left vs right column). Notice how nondeceptive episodes (middle) are much longer than deceptive episodes (bottom), whilst both are unsuccessful. Also note the absolute lack of search in the bedroom when changing wall colors deceptively.}
     \label{fig:qualitative_examples_decep_vs_nondecep}
 \end{figure}

\begin{figure*}[tbp]
     \includegraphics[trim=-0.1cm 0.5cm -0.1cm -0.5cm, clip, width=\textwidth]{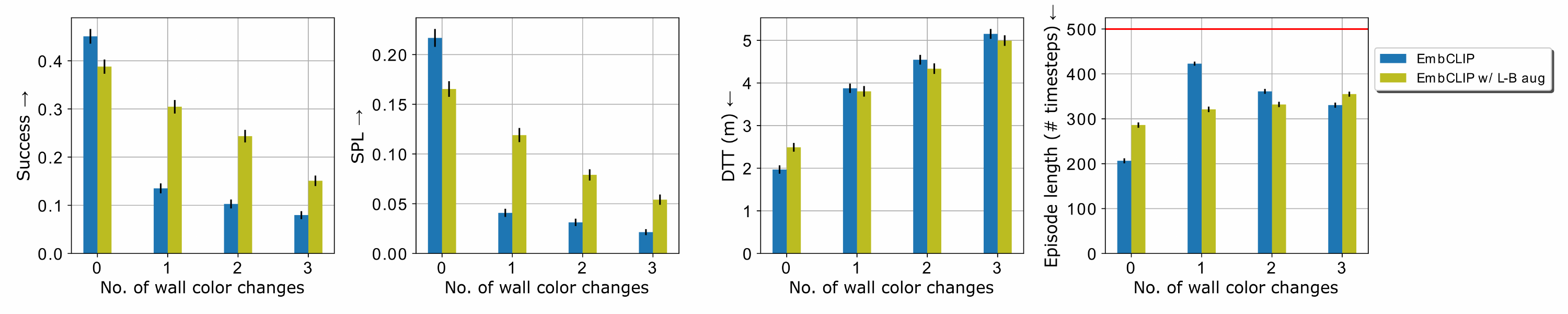}
     \caption{\textbf{Extending EmbCLIP \cite{EmbCLIP} with our L-B augmentations.} Performance on o.o.d. cases for EmbCLIP \cite{EmbCLIP} vs EmbCLIP with our L-B augmentations. Using our L-B augmentations, EmbCLIP generalizes better to scenes with different wall colors. }
     \label{fig:res_LB_aug_Objectnav}
\end{figure*}

How well does a SOTA ObjectNav method generalize to scenes with different wall colors and to what extent do shortcuts affect the o.o.d. generalization ability? We hypothesise the performance drop coheres with the number of wall color changes. Moreover, we posit a deceptive change will lead to more performance degradation than a nondeceptive change as the agent will be misled to search for the target object in the wrong room. Fig. \ref{fig:res_ood_gen_test} shows the performance. We report the mean over all episodes, the mean over episodes with deceptive changes and the mean over episodes with nondeceptive changes. We observe that changing the wall colors of only the target room already leads to a large decrease in performance. On average (blue mean bar), we observe a $67$\% relative drop in SPL (0.39 $\rightarrow$ 0.13) and $56$\% in success rate ($68$\% $\rightarrow$ $30$\%) going from 0 wall color changes to 1 wall color change, with even lower mean performance for more wall color changes. Indeed, we find that EmbCLIP generalizes poorly to scenes with different wall colors when using limited training data. Regarding deceptive vs nondeceptive changes, the Success, SPL and DTT metrics indicate that deceptive changes indeed deteriorate performance most, even more than multiple nondeceptive changes. Interestingly, however, we observe shorter episode lengths for deceptive changes. We conjecture that due to deceptive changes, the agent directly navigates towards the learned color of the target room, which is now placed in a non-target room, without exploring any other rooms. The agent will erroneously search this non-target room, but can not find the target object, and terminates the episode. In contrast, an agent will explore the entire scene when wall colors have only been changed nondeceptively, leading to longer episodes. We show a qualitative example of this behaviour in Fig. \ref{fig:qualitative_examples_decep_vs_nondecep}, where the agent has learned to look for a room with green walls instead of a bedroom. The agent is deceived by the green living room, and terminates the episode when it sees the sofa. This leads to a much shorter episode than in the nondeceptive example, where the agent explores large parts of the scene. Evidently, the agent has learned a shortcut strategy, it navigates towards a particular wall color instead of the right target room. 


\subsection{Benefit of Proposed L-B Augmentations}

Do our L-B augmentations make the agent's model more robust to shortcuts, i.e., more domain invariant against biased wall color and room type? We integrate our L-B augmentation method within the EmbCLIP architecture as a single extra layer, as detailed in Section \ref{meth:LB_aug}. Note that training time is only marginally longer: from 88 to 90 GPU-hours (both 30M steps). Fig. \ref{fig:res_LB_aug_Objectnav} shows a comparison of EmbCLIP with and without our L-B augmentations. EmbCLIP's performance already degrades significantly after changing wall colors of the target room (1 wall color change). We observe 69\% relative drop in success ($45\% \rightarrow 14\%$) and 82\% drop in SPL ($0.22 \rightarrow 0.04$). In contrast, our method shows improved domain generalization. When changing wall colors of the target room (1 wall color change), our method incurs only a 23\% relative drop in success ($39\% \rightarrow 30\%$) and 29\% drop in SPL ($0.17 \rightarrow 0.12$). We observe less performance degradation with an increasing number of wall color changes than EmbCLIP. In the qualitative example of Fig. \ref{fig:intro_fig}, the agent is now able to find the sofa even though it is not in a living room with blue walls (training). The agent finds the sofa successfully in a living room with green walls (not seen during training).  These results demonstrate that our L-B augmentations are an interesting direction to make RL agents more robust to biases, by only adding one additional layer to the agent's model. 

\section{Conclusion and limitations}
We evaluated how well a SOTA method for ObjectNav generalizes to scenes with different wall colors, and studied to what extent shortcut learning influences this o.o.d. generalization. We found that, when deliberately limiting training data, only changing wall colors in testing scenes decreases performance significantly, with the root cause being the deceptive wall color changes. We proposed Language-Based (L-B) augmentation to mitigate shortcut learning. By encoding text descriptions of variations of the dataset bias, and leveraging the multimodal embedding space of CLIP, we were able to augment agent's visual representations directly at feature-level. Finally, experiments showed that our L-B augmentation method is able to improve domain generalization to scenes with different wall colors in ObjectNav. When changing the target object's room, our method incurs a 23\% relative drop in success rate whilst the SOTA ObjectNav method's success rate drops 69\%.

To demonstrate the usefulness of our approach, we considered a simple case of shortcut bias e.g., wall color. Although our agents showed improved domain generalization, such simple cases may well be accommodated using conventional domain randomization methods in simulators which are easily modifiable. In future work, we hope to explore how to use natural language for augmentations addressing more intricate biases. For instance, bias at the object-level. Some objects might usually occur in combination with other objects (e.g., pillow on a bed). Agents could learn a bias for navigating towards the bedroom instead of also exploring the living room (e.g., pillow on a sofa). Moreover, as efficient ObjectNav requires agents to leverage useful semantic priors about the environment (e.g., object-room relations), it would be interesting to see how to use natural language to guide the exploration of agents in more unusual situations, where such priors are disadvantageous (e.g. pillow in the kitchen).  


\bibliographystyle{IEEEtran} 
\bibliography{IEEEabrv, IEEEexample}

\section*{Appendix}
\subsection{Target object selection}
\label{app:target_selection}
We select 3 target objects per room type. The selection is based on the object's overall frequency of occurrence in ProcTHOR-10k and if they have a clear semantic relation with one of the room types. We select different sized objects for each room type. Table \ref{tab:target_object_cats} shows an overview.

\subsection{Distribution evaluation episodes}
\label{App:eps_split}
We evenly distribute 1080 evaluation episodes over the 9 target objects, 5 scene layouts and possible wall color permutations for each test set. As there are more possible permutations for increasing number of wall color changes, the number of episodes per unique scene (combination of layout and wall color permutation) decreases. For example, for the 0-room test set, only 5 unique scenes are possible as only 1 wall color permutation is possible (no wall color changes w.r.t. training set). Hence, we run 216 episodes per unique scene. For the 1-room test set, we can change the target room to 2 different wall colors e.g., bedroom from green (train) to blue or red. In this case, we distribute the 1080 episodes over 10 unique scenes (5 layouts and 2 wall color permutations). We do the same for the 2-room and 3-room test set. Next, we evenly distribute over the 9 target objects. For instance, for the 0-room test set, we distribute the 216 episodes over the 9 target objects (24 per target object).

\subsection{Additional training details}
\label{App:train_hyperparams}
Table \ref{tab:training_params} details the hyperparameters we set for all of our training runs. We use DD-PPO \cite{DDPPO_PointNav} and Generalized Advantage Estimation (GAE) \cite{GAE}, parameterized by $\lambda = 0.95$.

\begin{table}[tbp]
    \centering
    \caption{\textbf{Action space description.} We use a 6-action discrete action space.}
        \begin{tabular}{lc}
            \toprule
            \textbf{Action} & \textbf{Description} \\ \hline
             MOVEAHEAD & \makecell[c]{Moves the agent forward (if possible) by sampling \\ from $\mathcal{N}(\mu = 0.25m, \sigma = 0.005m)$.} \\ \\
            \makecell[l]{ROTATELEFT \\ ROTATERIGHT} & \makecell[c]{Rotates the agent left or right by sampling from \\  $\mathcal{N}(\mu = 30\degree, \sigma = 0.5\degree)$} \\ \\
            \makecell[l]{LOOKUP \\ LOOKDOWN} & \makecell[c]{Tilt the camera of the agent upward or \\ downward by $30 \degree$} \\ \\
            DONE & Special action of the agent to terminate the episode. \\ \bottomrule
            \end{tabular}
        \label{tab:action_space}
\end{table}

\begin{table}[tbp] 
    \centering
    \caption{\textbf{Target objects selected for each room type.}}
        \begin{tabular}{lc}
            \toprule
            \textbf{Room type} & \textbf{Target object category} \\ \hline
            Kitchen & \makecell[c]{Fridge \\ Kettle \\ Apple} \\ \cdashline{1-2}
            Living room & \makecell[c]{Sofa \\ Television \\ Newspaper} \\ \cdashline{1-2}
            Bedroom & \makecell[c]{Bed \\ Dresser \\ Alarm clock} \\
            \bottomrule
            \end{tabular}
        \label{tab:target_object_cats}
\end{table}

\begin{table}[tbp] 
    \centering
    \caption{\textbf{Training hyperparameters.}}
        \begin{tabular}{lc}
            \toprule
            \textbf{Hyperparameter} & \textbf{Value} \\ \hline
            No. of GPUs & $2$ \\
            No. environments per GPU & $20$ \\
            Rollout length & $192$ \\
            No. mini-batches per rollout & $1$ \\            
            PPO epochs & $4$ \\
            Discount factor ($\gamma$) & $0.99$ \\
            GAE \cite{GAE} parameter ($\lambda$) & $0.95$ \\
            Value loss coefficient & $0.5$ \\
            Entropy loss coefficient & $0.01$ \\
            PPO clip parameter ($\epsilon$) & $0.1$ \\
            Gradient clip norm & $0.5$ \\
            Optimizer & Adam \\
            Learning rate & 3e-4 \\
            \bottomrule
            \end{tabular}
        \label{tab:training_params}
\end{table}

\end{document}